\documentclass[conference]{IEEEtran}

\usepackage{cite}
\usepackage{amsmath,amssymb,amsfonts}
\usepackage{algorithm, algpseudocode}
\usepackage{multicol}
\usepackage{multirow}
\usepackage{graphicx}
\usepackage{textcomp}
\usepackage{xcolor}

\begin{document}
\title{FSD: Fully-Specialized Detector via Neural Architecture Search\\
\thanks{Identify applicable funding agency here. If none, delete this.}
}

\author{\IEEEauthorblockN{Zhe Huang}
\IEEEauthorblockA{\textit{Department of Computer Sciences} \\
\textit{University of Wisconsin-Madison}\\
Madison, WI, USA \\
zhuang334@wisc.edu}
\and
\IEEEauthorblockN{Yudian Li}
\IEEEauthorblockA{\textit{Computer Science \& Engineering Department} \\
\textit{University of California, San Diego}\\
La Jolla, CA, USA \\
yudian.li.research@gmail.com}
}

\maketitle

\begin{abstract}
Most generic object detectors are mainly built for "standard" object detection tasks such as COCO\cite{lin2014microsoft} and PASCAL VOC\cite{Everingham10}. They might not work well and/or efficiently on tasks of other domains consisting of images that are visually different from standard datasets. To this end, many advances have been focused on adapting a general-purposed object detector with limited domain-specific designs. However, designing a successful task-specific detector requires extraneous manual experiments and parameter tuning through trial and error. In this paper, we first propose and examine a fully-automatic pipeline to design a fully-specialized detector (FSD) which mainly incorporates a neural-architectural-searched model by exploring ideal network structures over the backbone and task-specific head. On the DeepLesion dataset, extensive results show that FSD can achieve 3.1 mAP gain while using approximately 40\% fewer parameters on binary lesion detection task and improved the mAP by around 10\% on multi-type lesion detection task via our region-aware graph modeling compared with existing general-purposed medical lesion detection networks. 
\end{abstract}

\begin{IEEEkeywords}
    Object detection, neural architecture search, graph convolutional network.
\end{IEEEkeywords}

\section{Introduction}
Deep Neural Networks(DNNs) are widely used in a large variety of applications, such as image classification\cite{he2015deep},\cite{ma2018shufflenet},\cite{zhang2018shufflenet},\cite{wu2020mebow} and objective detection\cite{tian2019fcos},\cite{fasterrcnn},\cite{girshick2013rich} and achieved significant results. 

Most of these networks are general-purposed and are not designed for specific tasks to achieve optimal performances. This might not be a problem since images in most existing tasks are visually similar, featuring daily objects, natural scenes, etc. 

However, these advances in conventional object detection are mainly designed for natural images instead of CT scans. Among lesion detection tasks, lesions from CT scans are often similar to some non-lesion areas. Besides, lesions are usually varied in size and seriously class-imbalanced. Are those general-purposed networks well-adapted to the medical image domain? Specifically, for two-stage lesion detection tasks, will backbones and heads, which are fit for other task domains and data domains, be promising to bring superior performance on medical lesion scans that barely share any resemblance with other images. Thus, a generic object detector may face efficiency or accuracy issues.

Recent works proposed various methods to address such issues. \cite{shin2016deep} detected medical lesions by utilizing the off-the-shelf CNN with weights pretrained on ImageNet~\cite{alexnet}. Based on medical segmentation annotations, Paual et al.~\cite{jaeger2018retina} combined RetinaNet~\cite{lin2017focal} with U-Net~\cite{ronneberger2015u} to improve the medical detection performance. To benefit from 3D context, Yan et al.~\cite{yan20183d} aggregated multiple 2D images to incorporate 3D contextual information and used R-FCN~\cite{dai2016r} as the detection framework. 

However, these domain-specific designs require numerous attempts to modify the network and fine-tune the parameters, with no guarantee of finding an optimal solution on the targeted dataset. Also, in an effort to reduce such manual labor, they often only innovate a small portion of the full detector, i.e. building their customized models on top of a generic object detection network. Are their proposed solutions really the most efficient and the best working ones on non-standard tasks such as lesion detection?

To address the above-mentioned problem, we are motivated to design a fully-specialized detector directly built on the medical lesion detection domain. To fulfill this goal, it is intuitive to take advantage of existing techniques that are efficiently and effectively designed for detection tasks, such as the Faster-RCNN~\cite{fasterrcnn}, RetinaNet~\cite{lin2017focal}, FPN~\cite{lin2017feature}, RFBNet~\cite{Liu_2018_ECCV} and Relation Network~\cite{Hu_2018_CVPR}. However, to successfully incorporate those detection strategies, strong domain knowledge and numerous trials and errors for adjustments are required~\cite{wu2023learning}, which is prohibitive and troublesome. Thus, in an automated manner, we first propose a neural architecture search (NAS) pipeline on lesion detection, named FSD-NAS, with well-defined domain-specific search spaces by taking the characteristics of lesion images into consideration. Thanks to the progress of differentiable NAS~\cite{liu2019auto,liu2018darts}, we are able to specialize in a lesion detector with minimal human efforts and expertise.

First, we turn to NAS to automatically design a light-weighted and lesion-specialized backbone like MobileNet~\cite{howard2017mobilenets} in a proxyless way. We show that by using a domain-specialized backbone, lesion detection performance can be improved due to the fact that the quality of feature extraction for medical lesion slices is elevated. To avoid the overuse of over-parameterized FC layers and be more efficient, we obtain a task-specific convolutional head via NAS that explores dedicated detection operations in the search space. We seamlessly incorporate the lesion-specialized backbone with the lesion-specialized head in an end-to-end network for medical lesion detection, called FSD.  

Additionally, we propose a region-aware graph module to model the relativity of detection proposals, assisting in the multi-type lesion detection task. Specifically, based on the proposals of each lesion scan, we model a learned regional relationship to pay more attention to relevant contextual information across different regions, which is beneficial to accurately diagnose and provide a way for using interpretable visualization to assist radiologists in making judgments. 

The main contributions are three-fold.
1) To the best of our knowledge, we make the first attempt to exploit a neural architecture search (NAS) method for lesion detection by taking the characteristics of domain-specific images into consideration. Unlike classical NAS methods targeted at classification tasks, our automated pipeline is for detection tasks, searching backbones and heads from scratch in a proxyless way. 2) We propose a region-aware graph module to learn regional relationships and pay more attention to relevant contextual information across different regions, which can improve detection accuracy and provide interpretability to experts. 3) Extensive experiments evaluated on the large-scale DeepLesion dataset demonstrate the effectiveness of our specialized models and achieve state-of-the-art results on both binary lesion detection tasks and multi-type lesion detection tasks with many fewer parameters.

\section{Related Work}
\subsection{Neural Architecture Search}

Aimed to relieve human experts from the labor of designing efficient networks, the neural architecture search (NAS) is born to automatically search effective neural topologies with specific search spaces. 
NAS has attracted more and more attention and lots of NAS variants have been developed for searching, such as evolutionary methods~\cite{real2018regularized,miikkulainen2019evolving}, reinforcement learning methods~\cite{baker2016designing,zoph2016neural} and differentiable optimization based methods~\cite{liu2019auto,cai2018proxylessnas,zhao2021few,bender2018understanding}. Because NAS is capable of designing architectures that effectively achieve high performance, it is more and more important to employ NAS in various tasks, such as image classification~\cite{real2017large,real2018regularized}, semantic segmentation~\cite{liu2019auto,chen2018searching} and object detection~\cite{DBLP:journals/corr/abs-1904-07392}.
Recently, Liu et al.~\cite{liu2019auto} and Chen et al.~\cite{chen2018searching} proposed a semantic image segmentation framework that extends NAS beyond image classification to dense image prediction. Lin and Ghiasi et al.~\cite{DBLP:journals/corr/abs-1904-07392} proposed a NAS-FPN to expand NAS method to feature pyramid architecture for object detection. In this paper, we first attempt to investigate the NAS on the detection backbone and task-specific head for classification and localization, which discovers networks with better performance and fewer parameters compared with traditional detection networks.

\subsection{Lesion Detection}

Due to varied sizes and insignificant feature differences in comparison with other non-lesion parts, medical lesion detection is challenging. It is a labor-intensive task for radiologists to localize and classify all the lesions in the full CT scan space. Hopefully, medical lesion detection could assist them in getting the localization and identification of lesions done and reduce their workloads. Recent medical lesion detection works~\cite{yan20183d,yan2018deeplesion,du2018real} have been proposed to directly extend the existing detection framework designed for natural images instead of medical CT scans. Yan et al.~\cite{yan20183d} aggregated multiple 2D feature maps to incorporate 3D contextual information and simply used R-FCN~\cite{dai2016r} as a detection framework. Yan et al.~\cite{yan2018deeplesion} introduced a large-scale medical lesion dataset called DeepLesion, and modeled medical lesion detection via conventional detection frameworks. In this paper, we propose the NAS on lesion detection to search for a new detection architecture that better processes medical lesion data. Moreover, inspired by~\cite{fang2017object}, we present a region-aware graph module to enhance interpretability. Using these techniques, we achieve state-of-the-art performance on the DeepLesion dataset with fewer parameters compared with using conventional detection networks.

\begin{figure*}
	\centering
	\includegraphics[width=1\linewidth]{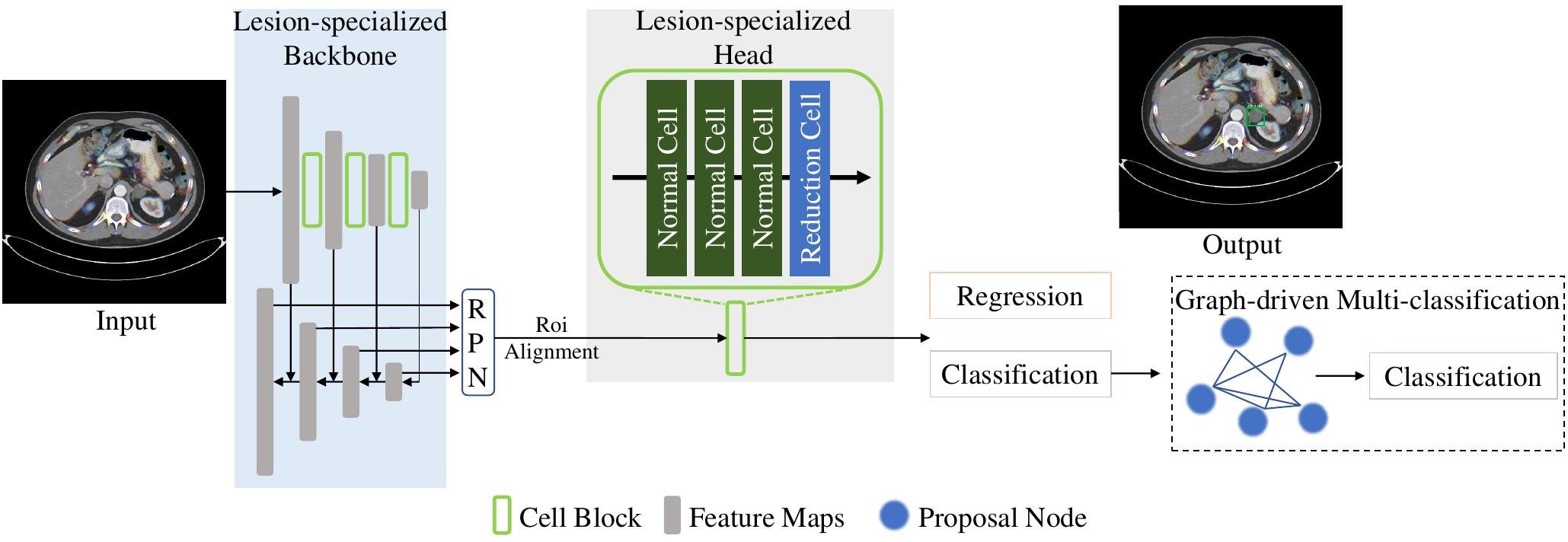}
	\vspace{-7mm}
	\caption{The overview of our specialized lesion detection network, which contains an FSD and a region-aware graph module.}
	\label{fig:framework}
	\vspace{-4mm}
\end{figure*}

\section{Methodology}

In this section, we introduce our pipeline of neural architecture search over the medical lesion detection domain (FSD-NAS) and our unique design of search spaces and search strategies. We then describe a graph reasoning method to further explore the design of the lesion-specialized network. The whole pipeline of our specialized medical lesion detection network can be seen in Figure~\ref{fig:framework}. 
\subsection{Objectives}

Following DARTS~\cite{liu2018darts}, we carry out our FSD-NAS in a differentiable way of joint optimization of the network architecture, denoted as $\alpha$, and parameter weights, denoted as $w$. Targeting to find the optimal architecture $\alpha^*$ with optimal weights $w^*$, we minimize the train loss and validation loss, denoted as $\mathcal{L}^{train}$ and $\mathcal{L}^{val}$, respectively. 

For modeling detection tasks, we separate the training loss $\mathcal{L}^{train}$ and the validation loss $\mathcal{L}^{val}$ into $\mathcal{L}^{train}_{bone}$ and $\mathcal{L}^{train}_{head}$, $\mathcal{L}^{val}_{bone}$ and $\mathcal{L}^{val}_{head}$, respectively, representing the corresponding loss of corresponding part (i.e. the backbone and the head) of the network. We further separate parameter weights $w$ into $w_{bone}$ and $w_{head}$, $\alpha$ into $\alpha_{bone}$ and $\alpha_{head}$, denoting parameter weights and the architecture encoding of the backbone and the detection head, respectively. In the following sections, we show our formulation of NAS problems using the above-mentioned symbols.

\subsection{Lesion-specialized Backbone (FSD Backbone)}

We can naturally divide a two-stage detection pipeline into two sequential parts, which are the feature extraction at first and regional classification \& localization thereafter. From a lesion specialization perspective, we would like to improve the feature representation of medical lesion images before selecting a region of interest and making regional detections. Thus, we flatten the nested joint optimization to only carry out NAS over the backbone at first, which is formulated as
\begin{equation}
\vspace{-2mm}
w^*_{bone} = \mathrm{argmin}_{{w}^{cell}_{bone}} \mathcal{L}^{train}(w, \alpha_{bone}),     
\end{equation}

\begin{equation}
\alpha^*_{bone} = \mathrm{argmin}_{\alpha_{bone}} \mathcal{L}^{val}(w^*_{bone}, \alpha_{bone}),     
\end{equation}

where $w^{cell}_{bone}$ represents parameter weights in search cells of the backbone, which can be a subset of $w_{bone}$ when there are other non-NAS parts in the backbone. In our case, we keep the architecture of the first two blocks of our backbone to be fixed. $\mathcal{L}^{train}(w, \alpha_{bone})$ means that we optimize all parameter weights in the detection pipeline (i.e. both backbone and head) to make the search process in an end-to-end manner. 

\textbf{FSD Backbone Search Space.} We carefully design the search space to be pragmatic and appropriate for the medical lesion detection task. While most of them are typical convolutional operations for building a general-purpose network, we add task-oriented or novel operations for specialization and remove some operations which appear to be less useful. First, we remove pooling mechanisms as they are generally considered harmful for the backbone of tiny object detection~\cite{scherer2010evaluation}. As medical lesions vary in magnitude and most are relatively small objects, pooling operations seem to be unnecessary and adverse for the feature representation. Moreover, we add a "res2conv"~\cite{gao2019res2net} operation to perform multi-scale fusion to accommodate the diverse scales of medical lesions. Specifically, our search space includes the following 8 operations:

\begin{tabular}{l l}
	\textbullet{\enskip conv 3x3 (dilation = 1)} & \textbullet{\enskip none} \\
	\textbullet{\enskip skip connection} & \textbullet{\enskip depthwise conv 3x3}  \\
	\textbullet{\enskip factorized conv 5x5} & \textbullet{\enskip res2conv 3x3} \\
	\textbullet{\enskip conv 3x3 (dilation = 3)} & \textbullet{\enskip conv 3x3 (dilation = 5)} \\
\end{tabular}

\subsection{Lesion-specialized Head (FSD Head)}

After searching for the backbone, we take one step further to search for a convolutional head for medical lesion detection to replace the overused fully-connected linear layers in the traditional detection head, which is designed mainly for common objects and multi-class classification. Different from the NAS on the backbone mentioned above, at the time we search for a suitable head customized for medical lesion detection, we already had the specialized backbone for lesion feature extraction. Hence, we can directly perform NAS on the detection head using pre-trained and fixed lesion-specialized FSD Backbone. We can formulate this optimization process as

\begin{align}
\min_{\alpha_{head}} \quad & \mathcal{L}^{val}_{head}(w^*_{head}) \label{eq:outer-head} \\
\text{s.t.} \enskip \quad & w^*_{head} = \mathrm{argmin}_{w_{head}} \mathcal{L}^{train}_{head}(w_{head}, \alpha_{head}). \label{eq:inner-head}
\end{align}

This can be interpreted as a NAS on lesion classification \& localization guided by the extracted feature of lesion detection task-oriented backbone. The algorithm of the FSD-NAS pipeline combining FSD Backbone \& FSD Head is shown in the Algo.~\ref{algo:pseudocode}.

However, since splitting a nested optimization into two less-dependent ones have the negative effect of parameter decoupling, searching results of the head could be sub-optimal. Besides, due to the limitation of searching cells, the original FC-based head is excluded from searching, meaning chances are the best convolutional head design paring with a specific backbone could sometimes still worse than the original FC head. Therefore, searching for task-specific heads is more with the intention to reduce the total number of parameters rather than performance-driven purposes. 

\textbf{FSD Head Search Space.} Apart from 8 operations from backbone searching, which we find them quite helpful, we add another five operations into our search space for lesion-specialized detection head, including the squeeze-and-excitation operation and the non-local block. Hence, the search space for FSD Head can be listed as: 

\begin{tabular}{l l}
 	\textbullet{\enskip none} & \textbullet{\enskip conv 3x3 (dilation = 1)} \\
	\textbullet{\enskip skip connection} &  \textbullet{\enskip depthwise conv 3x3} \\
	\textbullet{\enskip factorized conv 5x5} & \textbullet{\enskip depthwise conv 5x5} \\
	\textbullet{\enskip conv 3x3 (dilation = 3)} & \textbullet{\enskip conv 3x3 (dilation = 5)} \\
	\textbullet{\enskip avg pool 3x3} & \textbullet{\enskip res2conv 3x3} \\
	\textbullet{\enskip non-local} & \textbullet {\enskip max pool 3x3} \\
	\textbullet{\enskip squeeze-and-excitation}	
\end{tabular}

\begin{algorithm}[t]
        \begin{algorithmic} [1]
            \State Search for $\alpha^*_{bone}$ via proposed optimization and search space.
            \State Derive the FSD Backbone architecture based on the previous search result.
            \State Use FSD Backbone as the backbone, searching for $\alpha^*_{head}$ via proposed process and search space.
            \State Derive the FSD Head architecture based on the previous search result.
            \State Combine FSD Backbone \& FSD Head together as FSD to perform lesion detection.
        \end{algorithmic}
	\caption{{\sc FSD-NAS} -- FSD Backbone w/ FSD Head}
	\label{algo:pseudocode}
\end{algorithm}
\subsection{Region-aware Graph Module}
Given the object or region set $\mathcal{O}=\{o_i\}_{i=1}^N$ of an input image $\mathcal{I}$, we seek to construct an undirected graph $\mathcal{G}=\{\mathcal{V,E}\}$ over $\mathcal{O}$, where $\mathcal{V}=\{v_i\}_{i=1}^N$ is the node set and $\mathcal{E}=\{e_{ij}\}$ is the edge set.
Each node $v_i\in\{\mathcal{V}\}$ corresponds to a visual object $o_i \in \mathcal{O}$ and the associated feature vector with $d$ dimensions indicates $\mathbf{v}_i\in\mathbb{R}^d$. The $e_{ij}$ denotes the relationship between $o_i$ and $o_j$, and an adjacency weight matrix $\mathbf{A}$ is learned according to the edge connections denoted as $\{i,j,A_{ij}\}=\{e_{ij}\}\in\mathcal{E}$. By concatenating the joint embedding $\mathbf{v}_i$ together into a matrix $\mathbf{X}\in\mathbb{R}^{N\times d}$, we define the adjacency matrix for an undirected graph with self-loops as $\mathbf{A}=\mathbf{X}\mathbf{X}^T$. Here, we formulate the region-aware graph module as follows:
\begin{align}
& \mathbf{A_e} = \varphi_{\mathbf{w}}(\mathbf{A}^T)\phi_{\mathbf{w}}(\mathbf{A})\delta_{\mathbf{w}}(\mathbf{A}^T), \\ 
& \mathbf{X_e} = \sigma_{\mathbf{w}}(\mathbf{A_e}\mathbf{X}),
\end{align}  
where $\varphi_{\mathbf{w}},\phi_{\mathbf{w}},\delta_{\mathbf{w}}$ separately denote the learnable mapping matrix to map the value $A_{ij}$ into self-attention space for enhancing the correlations of each edge. Then given the enhanced adjacency weight matrix $\mathbf{A_e}$, a graph convolution operation~\cite{kipf2017semi} is used with a trainable matrix $\sigma_{\mathbf{w}}$ to propagate the relationship information to each node representation and update the graph.

When it comes to the implementation, our design can be seen as an add-on graph convolutional module that extends the original detection head. Taking a batch made up of 4 CT slices with 256 instances per slice as an example, after feature extraction by the detection head, we will end up having $N \times D$ shaped features after tensor squeezing, where $N$ denotes the number of all instances (1024 in this case) and $D$ denotes the dimension of the feature (e.g. 1024). We reshape it into $N_{s} \times N_{i} \times D$, where $N_{s}$ denotes the number of slices, which is 4, and $N_{i}$ denotes the number of instances in per slice, which is 256 here. In this way, our feature can be separated into $N_{s}$ parts with each part containing features of instances from the same slice. Then we feed this transformed feature into our graph module. First, a batch matrix-matrix production is performed with the input and its transpose to form a relational matrix sized as $N_{s} \times D \times D$. After that we perform several 1x1 convolutions to encode this matrix, find out relationships between instances in the same slice, and finally utilize this encoded relation to help the lesion classification. By graph modeling, our model successfully takes correlated information into consideration to boost performance.

\section{Experiments}	
In this session, we evaluate our proposed FSD on the large-scale DeepLesion dataset~\cite{yan2018deep} and achieve superior performance with fewer parameters for two different medical detection tasks (binary lesion detection and multi-class lesion detection). We demonstrate the effectiveness of each module in our network via ablation studies. 

\subsection{DeepLesion Dataset}
We carry out extensive experiments on DeepLesion~\cite{yan2018deep} dataset, a large-scale lesion detection benchmark dataset containing 32,120 CT key slices with 1-3 lesions annotated per slice, adding up to total of 32,735 lesions with bounding boxes. The neighboring slices of those key slices are also provided as optional 3D context information. This dataset is officially randomly split into a training set consisting of 22,496 lesions from 22,919 key slices, a validation set with 4,793 lesions from 4,889 key slices, and a test set made up of 4,831 lesions from the rest of 4,927 key slices. All the data from the official validation and test set are annotated with one of the seven lesion types, which are bone(BN), abdomen(AB), mediastinum(ME), liver(LV), lung(LU), kidney(KD), soft tissue(ST) and pelvis(PV) lesions, respectively, whereas training data are lacking in detailed lesion type information. Thus, the DeepLesion dataset can be used to perform binary lesion detection tasks as well as multi-type lesion detection studies thanks to the great variety of lesion types and provided annotations.

\subsection{Evaluation Criteria}
For a comprehensive evaluation of the detection results of our FSD, we adopt some common metrics in general object detection tasks, which are the mean average precision (mAP) and the overall mean recall rate~\cite{DBLP:journals/corr/LinMBHPRDZ14}. Moreover, following~\cite{yan20183d}, we also evaluate our method using the sensitivity under different false positives per image (FPPI) corresponding with different IoU thresholds and their intersection over the detected bounding-box (IoBB) thresholds counterparts. 

\subsection{Implementation Details}
We conduct all experiments using Pytorch~\cite{paszke2017automatic} with 8 GPUs.
For each key slice in the DeepLesion dataset, we concatenate neighboring 2 slices to form a 3-channel image and follow the official guideline~\footnote{https://nihcc.app.box.com/v/DeepLesion/folder/50715173939}. for data conversion, resulting in an 8-bit, 3-channel image with RGB values ranging from 0 to 255.
For all of our detection experiments, results are reported on the official test set. We adopt FPN~\cite{lin2017feature} and keep the same settings for all experiments, including our baseline. In all experiments, the shorter size of input images is set to 512. Anchor ratios are 0.5, 1, 2 and anchor scales are 2, 3, 4, 6, and 12.
For the binary lesion detection task, we set the initial learning rate to be 0.005 per sample and train our networks for 12 epochs with one additional warmup epoch at the very beginning of the training. The learning rate is multiplied by 0.1 when the training reaches the 8th and the 11th epoch and SGD optimizer with a momentum of 0.9 is used. 
For the multi-class lesion detection task, we set the initial learning rate to be 0.00125 per sample to stabilize the finetuning process and finetune for 15 epochs with two additional warmup epochs, using the same learning rate scheduler in the aforementioned binary detection.
For both NAS processes, we set the initial learning rate for network weights optimization to 0.01 per image with SGD optimizer (momentum set to 0.9 and weight decay set to 0.0003) and cosine learning scheduler annealing to 0.0001 gradually. The learning rate for the architecture search is set to be 0.0024 per image with Adam~\cite{kingma2014adam} optimizer with 0.001 weight decay.
Following~\cite{liu2018darts}, we split the training data into two equal halves for architecture searching and weight updating. We formulate our FSD Backbone and FSD Head by augmenting based on search cells.

\begin{table*}[]
	\small
	\centering
	\begin{tabular}{cccccc}
		\hline
		& \begin{tabular}[c]{@{}c@{}}3DCE~\cite{yan20183d} \\ w/ 3 slices\end{tabular} & \begin{tabular}[c]{@{}c@{}}3DCE~\cite{yan20183d} \\ w/ 27 slices\end{tabular} & \begin{tabular}[c]{@{}c@{}}FPN~\cite{lin2017feature} \\ (R50 + FC Head)\end{tabular} & \begin{tabular}[c]{@{}c@{}}FSD Backbone \\ w/ FC Head\end{tabular} & \begin{tabular}[c]{@{}c@{}}FSD Backbone\\ w/ FSD Head\end{tabular} \\ \hline
		mAP@{[}.5:.95{]} & 25.6 & 29.2 & 35.3 & \textbf{38.5} & 38.4 \\ 
		
		\# of parameters & 17.2 & 25.2 & 39.5 & 37.8 & \textbf{24.7} \\ \hline
	\end{tabular}
	\smallskip
	\caption{Results of mAP@{[}.5:.95{]} and the number of parameters in millions in the network. IoU is used to compute overlapping area. Compared with other state-of-the-art medical detection methods on the DeepLesion dataset, our mAP results on its binary detection task are better. Specifically, by lesion-specialized design (FSD Backbone w/ FSD Head), we achieve 3.1 mAP gain while using 14.8M parameters less than our FPN baseline (ResNet-50 + FC Head).}
	\label{tab:binary-precision}
	\vspace{-2mm}
\end{table*}

\begin{table*}[]
	\small
	\centering
	\begin{tabular}{c|c|c|cccccc}
		\hline
		\multicolumn{3}{c}{\multirow{2}{*}{}} & \multicolumn{6}{c}{FPs per image (FPPI)} \\
		\multicolumn{3}{c}{} & 0.5 & 1 & 2 & 4 & 8 & 16 \\ \hline
		\multirow{2}{*}{\begin{tabular}[c]{@{}c@{}}3DCE~\cite{yan20183d}\\ w/ 3 slices\end{tabular}} & \multirow{10}{*}{\rotatebox[origin=c]{90}{Intersection Criteria}} & IoU & 56.49 & 67.65 & 76.89 & 82.76 & 87.03 & 89.82 \\
		&  & IoBB & 58.43 & 70.95 & 80.64 & 87.30 & 92.37 & 94.33 \\ \cline{1-1} \cline{3-9} 
		\multirow{2}{*}{\begin{tabular}[c]{@{}c@{}}3DCE~\cite{yan20183d}\\ w/ 27 slices\end{tabular}} &  & IoU & 62.48 & 73.37 & 80.70 & 85.65 & 89.09 & 91.06 \\
		&  & IoBB & 64.01 & 75.69 & 83.71 & 87.52 & 92.85 & 95.64 \\ \cline{1-1} \cline{3-9} 
		\multirow{2}{*}{\begin{tabular}[c]{@{}c@{}}FPN~\cite{lin2017feature}\\ (R50 + FC Head)\end{tabular}} &  & IoU & 66.21 & 75.18 & 81.62 & 86.73 & 90.57 & 93.36 \\
		&  & IoBB & 68.16 & 76.90 & 83.45 & 88.56 & 92.57 & 95.30 \\ \cline{1-1} \cline{3-9} 
		\multirow{2}{*}{\begin{tabular}[c]{@{}c@{}}FSD Backbone\\ w/ FC Head\end{tabular}} &  & IoU & 69.01 & 76.60 & 83.57 & 89.07 & 93.06 & 95.42 \\
		&  & IoBB & 71.38 & 79.56 & 86.36 & 91.41 & 95.03 & 97.09 \\ \cline{1-1} \cline{3-9} 
		\multirow{2}{*}{\begin{tabular}[c]{@{}c@{}}FSD Backbone\\ w/ FSD Head\end{tabular}} &  & IoU & 68.20 & 76.45 & 83.18 & 88.25 & 92.00 & 94.77 \\
		&  & IoBB & 70.42 & 78.91 & 85.73 & 90.78 & 94.44 & 96.91 \\ \hline
	\end{tabular}
	\smallskip
	\caption{Results of sensitivity at different false positives (FPs) per image, mAP@{[}.5:.95{]} in the network. Both IoU \& IoBB are used to compute overlapping area. We consistently outperform 3DCE and FPN baseline on the DeepLesion dataset. Although FSD w/ FC Head is slightly better than FSD w/ FSD Head, the latter uses 13.1M less parameters in total.}
	\label{tab:binary-sensitivity}
	\vspace{-5mm}
\end{table*}

\begin{figure*}
    \centering
    \includegraphics[width=1.0\linewidth]{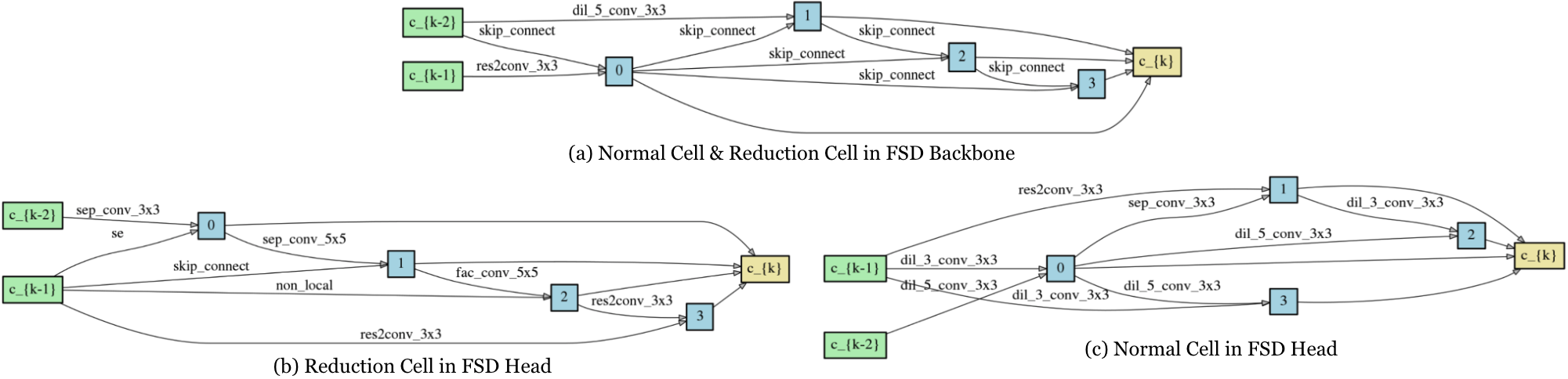}
    \caption{The searching results of our FSD Backbone and FSD Head.}
    \label{fig:cell}
\end{figure*}

\subsection{Results and Comparisons}

During this section, we show our state-of-the-art results on the DeepLesion dataset w.r.t. both binary and multi-class lesion detection, in comparason with others such as general-purposed FPN~\cite{lin2017feature} and DeepLesion-specialized 3DCE~\cite{yan20183d}, through measurements either common for detection tasks or specially designed for medical detection purposes (e.g. IoBB metrics).

\textbf{FSD Network Structure.} Please refer to Fig.~\ref{fig:cell}.

\begin{figure*}
	\centering
	\includegraphics[width=0.7\linewidth]{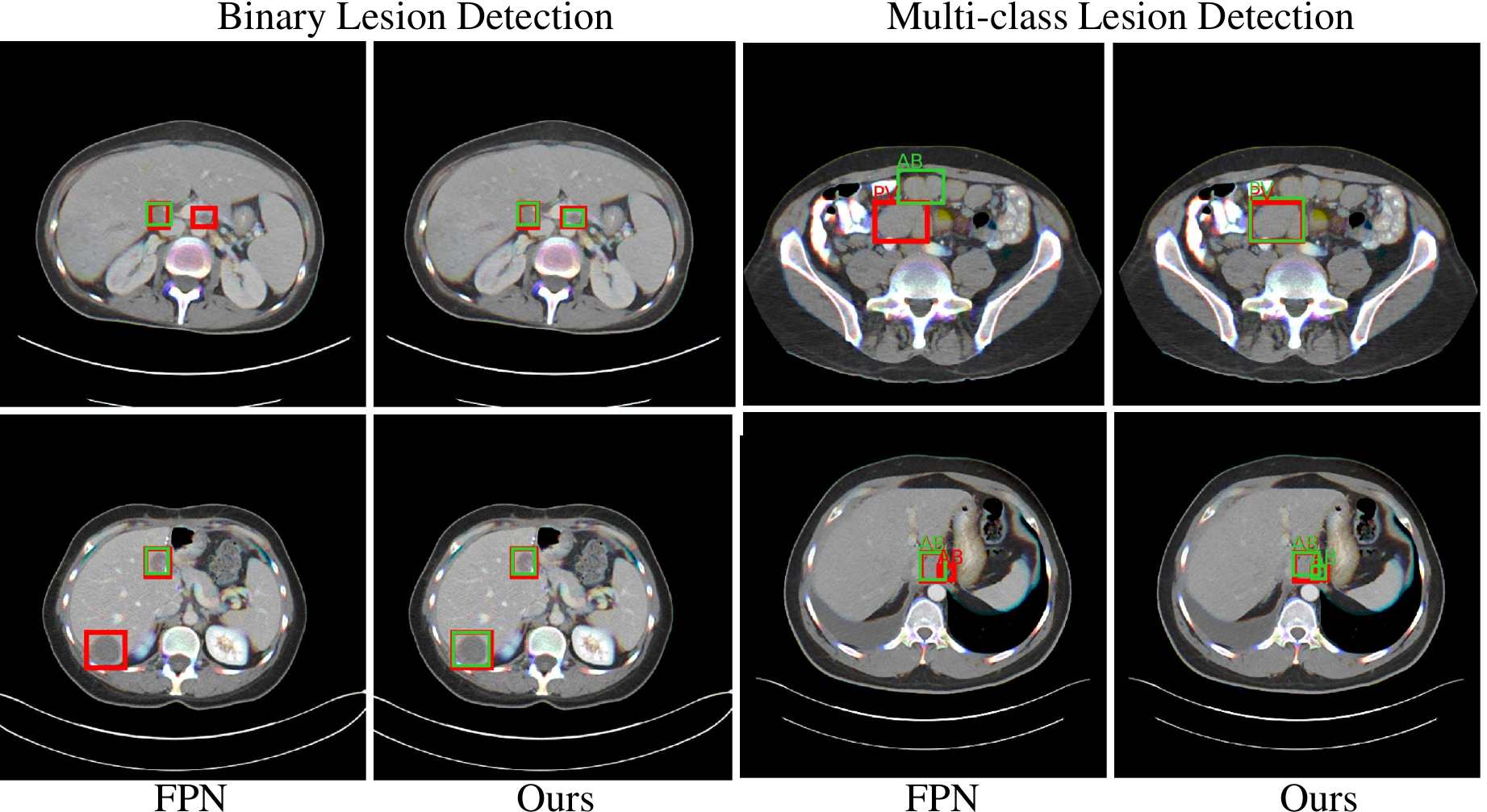}
	\vspace{-2mm}
	\caption{Qualitative results of binary and multi-class lesion detection. {\color{red}Red} boxes denote ground truth (GT) and {\color{green}green} boxes indicate prediction results. Those translucent white boxes denote predictions with relatively low confidence, which can be simply ignored. As is shown in many cases, our networks detect both small and large lesions well along with highly accurate location predictions.}
	\label{fig:result}
\end{figure*}

\textbf{Binary Lesion Detection.} For a comprehensive and thorough analysis, we report results of FPN, original 3DCE with 3 or 27 slices, FSD w/o FSD Head (FSD Backbone + FC head) and FSD (FSD Backbone + FSD Head) altogether with mAP and size comparison in Table~\ref{tab:binary-precision}. Not only do we surpass 3DCE~\cite{yan20183d} by 9.2\% in terms of mAP, compared with the conventional FPN, our specialized detector improves 3.1 mAP with around 40\% fewer parameters (38.4\% vs. 35.2\%, 24.7M vs. 39.5M), showing the effectiveness of our design.
Note that additional data (i.e. other surrounding CT slices in the DeepLesion dataset) serving as a 3D context is used in~\cite{yan20183d} to improve the lesion detection, whereas we do not use those additional data. 
As for our FSD, we only require standard 3-channel images as input, achieving highly-efficient feature representation and great performance improvement over all metrics. 
We also report sensitivity measurements in Table~\ref{tab:binary-sensitivity}. We notice that our method performs better at all FPPIs, especially at 0.5 FPPI compared to 3DCE (56.49 vs. 69.01 over IoU, 58.43 vs. 71.38 over IoBB), indicating our networks detect well both on annotated lesions or on missing-labeled lesions, which further demonstrates that our customized networks fit the need of medical lesion detection.
We also provide qualitative results, in Figure ~\ref{fig:result}, showing the high detection accuracy of our proposed FSD.

\begin{table*}[]
	\small
	\centering
	\begin{tabular}{c|c|cccccccc|c}
		\hline
		\multicolumn{2}{c|}{\multirow{2}{*}{}} & \multicolumn{8}{c|}{Class AP@{[}.5:.95{]}} & \\
		\multicolumn{2}{c|}{} & BN & AB & ME & LV & LU & KD & ST & \multicolumn{1}{c|}{PV} & mAP \\ \hline
		\multirow{2}{*}{\begin{tabular}[c]{@{}c@{}}FPN \\ w/o Graph\end{tabular}} & IoU & 24.03 & 18.99 & 27.82 & 31.36 & 38.18 & 15.62 & 21.75 & 18.19 & 24.5 \\
		& IoBB & 34.42 & 28.61 & 38.27 & 41.85 & 50.60 & 19.70 & 27.71 & 29.29 & 33.8 \\ \hline
		\multirow{2}{*}{\begin{tabular}[c]{@{}c@{}}FPN \\ w/ Graph\end{tabular}} & IoU & 17.08 & 19.06 & 30.44 & 27.45 & 40.24 & 22.13 & 20.35 & 22.21 & 24.9 \\
		& IoBB & 27.45 & 28.60 & 41.65 & 39.27 & 52.70 & 27.53 & 25.49 & 34.56 & 34.7 \\ \hline
		\multirow{2}{*}{\begin{tabular}[c]{@{}c@{}}FSD\\ w/o Graph\end{tabular}} & IoU & 19.42 & 22.05 & 30.02 & 31.73 & 44.31 & 10.64 & 22.52 & 21.76 & 25.3 \\
		& IoBB & 31.09 & 32.11 & 42.03 & 44.99 & 58.22 & 14.42 & 28.32 & 39.28 & 36.3 \\ \hline
		\multirow{2}{*}{\begin{tabular}[c]{@{}c@{}}FSD\\ w Graph\end{tabular}} & IoU & 23.75 & 21.81 & 28.95 & 30.73 & 44.15 & 20.99 & 24.06 & 23.24 & \textbf{27.7} \\
		& IoBB & 33.63 & 32.49 & 40.91 & 44.05 & 58.10 & 27.85 & 30.98 & 40.41 & \textbf{38.5} \\ \hline
	\end{tabular}
	\smallskip
	\caption{Results of class AP@[.5:.95] and mAP@[.5:.95] over all 4 different networks. Both IoU and IoBB are used to compute overlapping area. As is shown clearly in the table, mAPs, measured by both IoU and IoBB, are consistently increasing from FPN to FSD and from non-graph network to network with graph modeling, showing the effectiveness of our graph design.}
	\label{tab:multiclass-precision}
\end{table*}

\begin{table*}[!t]
	\vspace{-2mm}
	\small
	\centering
	\begin{tabular}{c|c|cccccccc|c}
		\hline
		\multicolumn{2}{c|}{\multirow{2}{*}{}} & \multicolumn{8}{c|}{Sensitivity (\%) at 4 FPPI} &  \\
		\multicolumn{2}{c|}{} & BN & AB & ME & LV & LU & KD & ST & \multicolumn{1}{c|}{PV} & Recall \\ \hline
		\multirow{2}{*}{\begin{tabular}[c]{@{}c@{}}FPN \\ w/o Graph\end{tabular}} & IoU & 86.11 & 81.33 & 89.35 & 93.43 & 88.64 & 86.32 & 78.37 & 86.31 & 51.7 \\
		& IoBB & 88.89 & 85.85 & 92.94 & 95.57 & 92.09 & 90.17 & 84.96 & 91.93 & 79.2 \\ \hline
		\multirow{2}{*}{\begin{tabular}[c]{@{}c@{}}FPN \\ w/ Graph\end{tabular}} & IoU & 79.63 & 87.89 & 92.01 & 92.86 & 92.73 & 90.17 & 85.84 & 88.26 & \textbf{55.5} \\
		& IoBB & 80.56 & 89.86 & 94.33 & 95.29 & 93.91 & 93.16 & 89.09 & 92.67 & 80.1 \\ \hline
		\multirow{2}{*}{\begin{tabular}[c]{@{}c@{}}FSD\\ w/o Graph\end{tabular}} & IoU & 86.11 & 84.40 & 91.20 & 91.43 & 91.55 & 84.62 & 83.19 & 87.29 & 52.2 \\
		& IoBB & 89.81 & 89.09 & 93.63 & 94.29 & 93.64 & 89.32 & 85.84 & 93.40 & 80.9 \\ \hline
		\multirow{2}{*}{\begin{tabular}[c]{@{}c@{}}FSD\\ w Graph\end{tabular}} & IoU & 80.56 & 87.98 & 92.59 & 93.86 & 92.36 & 88.03 & 85.25 & 88.02 & 54.4 \\
		& IoBB & 87.04 & 90.45 & 95.95 & 96.86 & 94.73 & 95.30 & 91.45 & 93.89 & \textbf{90.4} \\ \hline
	\end{tabular}
	\vspace{1mm}
	\caption{Results of sensitivity at 4 false positives (FPs) per image and recall@{[}.5:.95{]} over all classes. Both IoU \& IoBB are used to compute overlapping areas. As is shown above, the majority of lesion subtypes benefit from our FSD design and graph modeling in terms of sensitivity \& recall measured by both IoU \& IoBB criteria. Specifically, the overall recall under IoBB of our model, FSD w/ Graph, achieves 90.4\%, which is highly reliable for real-world lesion detection.}
	\label{tab:multiclass-recall}
	\vspace{-5mm}
\end{table*}

\begin{figure*}[!t]
	\centering
	\includegraphics[width=0.7\linewidth]{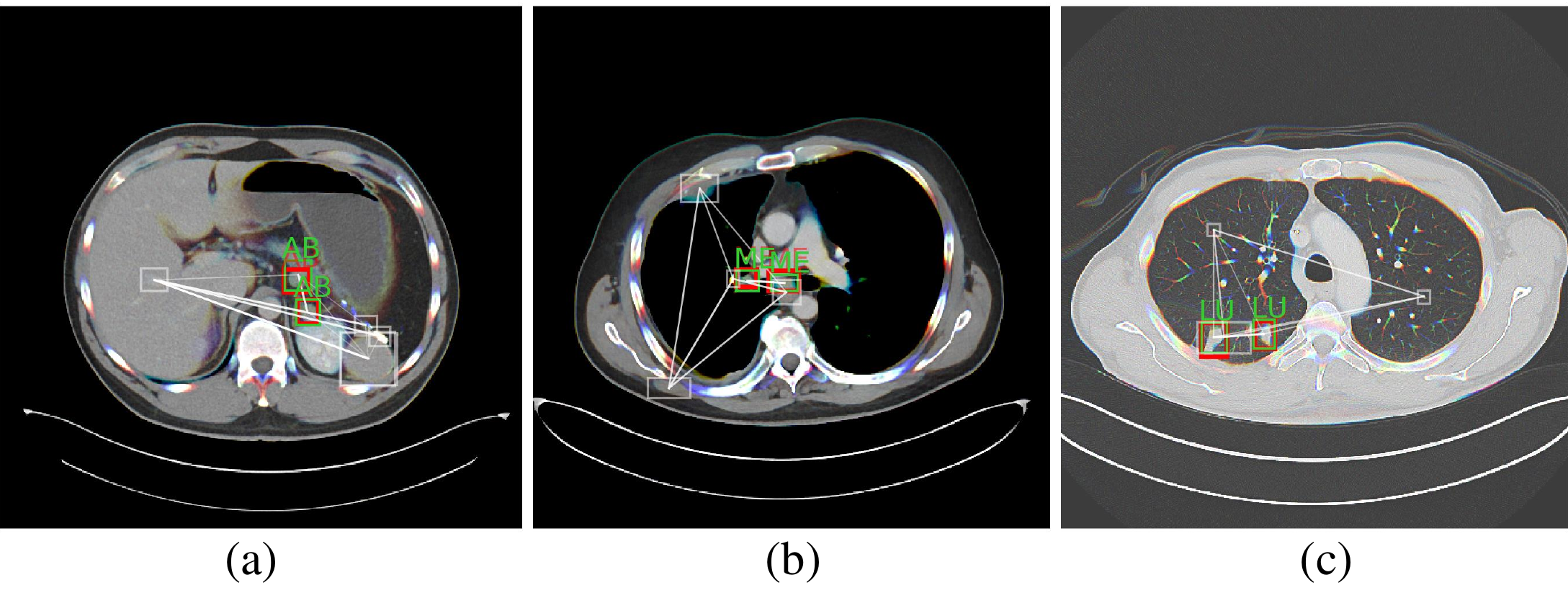}
	\vspace{-3mm}
	\caption{The qualitative results of the correlation between relevant lesion areas. {\color{red}Red} boxes denote ground truth (GT) and {\color{green}green} boxes indicate prediction results. Those white boxes denote predictions with relatively low confidence or background class. The opacity of lines between each pair of boxes denotes the strength of the correlation between them. It can be seen from the picture that different kinds of predictions are more intra-correlated as well as less inter-correlated.}
	\label{fig:graph}
\end{figure*}

\textbf{Multi-class Lesion Detection.} To further extend the scope of the specialization and customization of the network over lesion detection problems, we dive into the multi-class lesion detection problem, which is a relatively novel task within the scope of medical lesion detection and can better help radiologists. 

Due to the high similarity among different subtypes of lesions, we explore a graph reasoning approach to tackle those issues. In Table~\ref{tab:multiclass-precision}, the graph-based methods can significantly achieve performance gain over both IoU \& IoBB criteria, improving 0.4 mAP on FPN w/ Graph and 2.5 mAP on our FSD w/ Graph compared with their counterparts which are without graph. The fact that our task-specific FSD outperforms the FPN baseline (23.24\% vs. 22.21\% in terms of ) indicates that our lesion-specialized network design is well-suitable for discriminative lesion feature representation.
Moreover, we report results of sensitivity and recall@[.5:.95] over both IoU \& IoBB in Table~\ref{tab:multiclass-recall}. We show the overall recall smoothly increases from 51.7\% to 90.4\%, almost double due to the effect of the regional attention graph.
It confirms the idea that the proposals from the same CT scan are related negatively or positively and this type of information can be taken into consideration to perform fine-grained lesion detection with a small amount of annotated data. Qualitative comparison between FSD w/ and w/o Graph are shown in Figure ~\ref{fig:result}.

\subsection{Interpretability}
We offer the interpretation of our graph module via the visualization of our results of FSD w/ Graph, aiming for radiological reasoning. 
By adding the region-aware graph modeling after the multi-class classifier, we build a relational matrix over their classification features. 
We then use convolution layers to encode this relationship, assisting in multi-type lesion detection. To be specific, we visualize our graph modeling as is shown in Figure \ref{fig:graph}. For denoting the relationship between relevant lesion areas, we use white lines with different levels of opacity, from high to low, to indicate the strength of positive correlation between two predictions from highly positive-related to relatively non-positively correlated, correspondingly. From Figure \ref{fig:graph}, we can clearly see that true positives on the same slice are highly positively correlated and are dependent when the network makes its predictions, whereas false positives or backgrounds are related together. Those highly-scored true positives are very weakly associated with backgrounds or low-confident predictions, which is plausible as their features are presumably dissimilar to a large extent. In conclusion, our region-aware graph modeling is an effective way to exploit correlation with different proposals, improving lesion detection with sub-types in an interpretable manner.

\section{Conclusion}

In this paper, we presented a novel method for designing task-specific object detectors, especially for domains with visuals that differ substantially from conventional datasets. In order to determine optimal network structures for both the backbone and task-specific head, we proposed an automatic domain adaptation strategy via neural architectural search (NAS), dubbed FSD-NAS. We demonstrated that on DeepLesion dataset~\cite{yan2018deeplesion}, our model, fully-specialized detector (FSD), not only outperforms other state-of-the-art methods, but also achieves comparable results with reduced complexity. The findings suggest great potentials of this approach in domain-specific object detection tasks, providing a valuable avenue for future research and developments in the field.

\vfill

\bibliographystyle{IEEEtran}
\bibliography{egbib}

\begin{thebibliography}{10}
\providecommand{\url}[1]{#1}
\csname url@samestyle\endcsname
\providecommand{\newblock}{\relax}
\providecommand{\bibinfo}[2]{#2}
\providecommand{\BIBentrySTDinterwordspacing}{\spaceskip=0pt\relax}
\providecommand{\BIBentryALTinterwordstretchfactor}{4}
\providecommand{\BIBentryALTinterwordspacing}{\spaceskip=\fontdimen2\font plus
\BIBentryALTinterwordstretchfactor\fontdimen3\font minus
  \fontdimen4\font\relax}
\providecommand{\BIBforeignlanguage}[2]{{%
\expandafter\ifx\csname l@#1\endcsname\relax
\typeout{** WARNING: IEEEtran.bst: No hyphenation pattern has been}%
\typeout{** loaded for the language `#1'. Using the pattern for}%
\typeout{** the default language instead.}%
\else
\language=\csname l@#1\endcsname
\fi
#2}}
\providecommand{\BIBdecl}{\relax}
\BIBdecl

\bibitem{lin2014microsoft}
T.-Y. Lin, M.~Maire, S.~Belongie, J.~Hays, P.~Perona, D.~Ramanan,
  P.~Doll{\'a}r, and C.~L. Zitnick, ``Microsoft coco: Common objects in
  context,'' in \emph{Computer Vision--ECCV 2014: 13th European Conference,
  Zurich, Switzerland, September 6-12, 2014, Proceedings, Part V 13}.\hskip 1em
  plus 0.5em minus 0.4em\relax Springer, 2014, pp. 740--755.

\bibitem{Everingham10}
M.~Everingham, L.~Van~Gool, C.~K.~I. Williams, J.~Winn, and A.~Zisserman, ``The
  pascal visual object classes (voc) challenge,'' \emph{International Journal
  of Computer Vision}, vol.~88, no.~2, pp. 303--338, Jun. 2010.

\bibitem{he2015deep}
K.~He, X.~Zhang, S.~Ren, and J.~Sun, ``Deep residual learning for image
  recognition,'' in \emph{CVPR}, 2016.

\bibitem{ma2018shufflenet}
N.~Ma, X.~Zhang, H.-T. Zheng, and J.~Sun, ``Shufflenet v2: Practical guidelines
  for efficient cnn architecture design,'' in \emph{Proceedings of the European
  conference on computer vision (ECCV)}, 2018, pp. 116--131.

\bibitem{zhang2018shufflenet}
X.~Zhang, X.~Zhou, M.~Lin, and J.~Sun, ``Shufflenet: An extremely efficient
  convolutional neural network for mobile devices,'' in \emph{Proceedings of
  the IEEE conference on computer vision and pattern recognition}, 2018, pp.
  6848--6856.

\bibitem{wu2020mebow}
\BIBentryALTinterwordspacing
C.~Wu, Y.~Chen, J.~Luo, C.-C. Su, A.~Dawane, B.~Hanzra, Z.~Deng, B.~Liu, J.~Z.
  Wang, and C.-h. Kuo, ``Mebow: Monocular estimation of body orientation in the
  wild,'' \emph{2020 IEEE/CVF Conference on Computer Vision and Pattern
  Recognition (CVPR)}, Jun 2020. [Online]. Available:
  \url{http://dx.doi.org/10.1109/CVPR42600.2020.00351}
\BIBentrySTDinterwordspacing

\bibitem{tian2019fcos}
Z.~Tian, C.~Shen, H.~Chen, and T.~He, ``Fcos: Fully convolutional one-stage
  object detection,'' 2019.

\bibitem{fasterrcnn}
S.~Ren, K.~He, R.~Girshick, and J.~Sun, ``Faster r-cnn: Towards real-time
  object detection with region proposal networks,'' in \emph{NIPS}, 2015.

\bibitem{girshick2013rich}
R.~Girshick, J.~Donahue, T.~Darrell, and J.~Malik, ``Rich feature hierarchies
  for accurate object detection and semantic segmentation,'' \emph{arXiv
  preprint arXiv:1311.2524}, 2013.

\bibitem{shin2016deep}
H.-C. Shin, H.~R. Roth, M.~Gao, L.~Lu, Z.~Xu, I.~Nogues, J.~Yao, D.~Mollura,
  and R.~M. Summers, ``Deep convolutional neural networks for computer-aided
  detection: Cnn architectures, dataset characteristics and transfer
  learning,'' \emph{IEEE transactions on medical imaging}, vol.~35, no.~5, pp.
  1285--1298, 2016.

\bibitem{alexnet}
A.~Krizhevsky, I.~Sutskever, and G.~E. Hinton, ``Imagenet classification with
  deep convolutional neural networks,'' in \emph{Advances in Neural Information
  Processing Systems}, 2012.

\bibitem{jaeger2018retina}
P.~F. Jaeger, S.~A. Kohl, S.~Bickelhaupt, F.~Isensee, T.~A. Kuder, H.-P.
  Schlemmer, and K.~H. Maier-Hein, ``Retina u-net: Embarrassingly simple
  exploitation of segmentation supervision for medical object detection,''
  \emph{arXiv preprint arXiv:1811.08661}, 2018.

\bibitem{lin2017focal}
T.-Y. Lin, P.~Goyal, R.~Girshick, K.~He, and P.~Doll{\'a}r, ``Focal loss for
  dense object detection,'' in \emph{Proceedings of the IEEE international
  conference on computer vision}, 2017, pp. 2980--2988.

\bibitem{ronneberger2015u}
O.~Ronneberger, P.~Fischer, and T.~Brox, ``U-net: Convolutional networks for
  biomedical image segmentation,'' in \emph{International Conference on Medical
  image computing and computer-assisted intervention}.\hskip 1em plus 0.5em
  minus 0.4em\relax Springer, 2015, pp. 234--241.

\bibitem{yan20183d}
K.~Yan, M.~Bagheri, and R.~M. Summers, ``3d context enhanced region-based
  convolutional neural network for end-to-end lesion detection,'' in
  \emph{International Conference on Medical Image Computing and
  Computer-Assisted Intervention}.\hskip 1em plus 0.5em minus 0.4em\relax
  Springer, 2018, pp. 511--519.

\bibitem{dai2016r}
J.~Dai, Y.~Li, K.~He, and J.~Sun, ``R-fcn: Object detection via region-based
  fully convolutional networks,'' in \emph{Advances in neural information
  processing systems}, 2016, pp. 379--387.

\bibitem{lin2017feature}
T.-Y. Lin, P.~Doll{\'a}r, R.~Girshick, K.~He, B.~Hariharan, and S.~Belongie,
  ``Feature pyramid networks for object detection,'' in \emph{Proceedings of
  the IEEE Conference on Computer Vision and Pattern Recognition}, 2017, pp.
  2117--2125.

\bibitem{Liu_2018_ECCV}
S.~Liu, D.~Huang, and a.~Wang, ``Receptive field block net for accurate and
  fast object detection,'' in \emph{The European Conference on Computer Vision
  (ECCV)}, September 2018.

\bibitem{Hu_2018_CVPR}
H.~Hu, J.~Gu, Z.~Zhang, J.~Dai, and Y.~Wei, ``Relation networks for object
  detection,'' in \emph{The IEEE Conference on Computer Vision and Pattern
  Recognition (CVPR)}, June 2018.

\bibitem{wu2023learning}
C.~Wu, Y.~Pan, Y.~Li, and J.~Z. Wang, ``Learning to adapt to online streams
  with distribution shifts,'' \emph{arXiv preprint arXiv:2303.01630}, 2023.

\bibitem{liu2019auto}
C.~Liu, L.-C. Chen, F.~Schroff, H.~Adam, W.~Hua, A.~Yuille, and L.~Fei-Fei,
  ``Auto-deeplab: Hierarchical neural architecture search for semantic image
  segmentation,'' \emph{arXiv preprint arXiv:1901.02985}, 2019.

\bibitem{liu2018darts}
H.~Liu, K.~Simonyan, and Y.~Yang, ``Darts: Differentiable architecture
  search,'' \emph{arXiv preprint arXiv:1806.09055}, 2018.

\bibitem{howard2017mobilenets}
A.~G. Howard, M.~Zhu, B.~Chen, D.~Kalenichenko, W.~Wang, T.~Weyand,
  M.~Andreetto, and H.~Adam, ``Mobilenets: Efficient convolutional neural
  networks for mobile vision applications,'' \emph{arXiv preprint
  arXiv:1704.04861}, 2017.

\bibitem{real2018regularized}
E.~Real, A.~Aggarwal, Y.~Huang, and Q.~V. Le, ``Regularized evolution for image
  classifier architecture search,'' \emph{arXiv preprint arXiv:1802.01548},
  2018.

\bibitem{miikkulainen2019evolving}
R.~Miikkulainen, J.~Liang, E.~Meyerson, A.~Rawal, D.~Fink, O.~Francon, B.~Raju,
  H.~Shahrzad, A.~Navruzyan, N.~Duffy \emph{et~al.}, ``Evolving deep neural
  networks,'' in \emph{Artificial Intelligence in the Age of Neural Networks
  and Brain Computing}.\hskip 1em plus 0.5em minus 0.4em\relax Elsevier, 2019,
  pp. 293--312.

\bibitem{baker2016designing}
B.~Baker, O.~Gupta, N.~Naik, and R.~Raskar, ``Designing neural network
  architectures using reinforcement learning,'' \emph{arXiv preprint
  arXiv:1611.02167}, 2016.

\bibitem{zoph2016neural}
B.~Zoph and Q.~V. Le, ``Neural architecture search with reinforcement
  learning,'' \emph{arXiv preprint arXiv:1611.01578}, 2016.

\bibitem{cai2018proxylessnas}
H.~Cai, L.~Zhu, and S.~Han, ``Proxylessnas: Direct neural architecture search
  on target task and hardware,'' \emph{arXiv preprint arXiv:1812.00332}, 2018.

\bibitem{zhao2021few}
Y.~Zhao, L.~Wang, Y.~Tian, R.~Fonseca, and T.~Guo, ``Few-shot neural
  architecture search,'' in \emph{International Conference on Machine
  Learning}.\hskip 1em plus 0.5em minus 0.4em\relax PMLR, 2021, pp.
  12\,707--12\,718.

\bibitem{bender2018understanding}
G.~Bender, P.-J. Kindermans, B.~Zoph, V.~Vasudevan, and Q.~Le, ``Understanding
  and simplifying one-shot architecture search,'' in \emph{International
  conference on machine learning}.\hskip 1em plus 0.5em minus 0.4em\relax PMLR,
  2018, pp. 550--559.

\bibitem{real2017large}
E.~Real, S.~Moore, A.~Selle, S.~Saxena, Y.~L. Suematsu, J.~Tan, Q.~V. Le, and
  A.~Kurakin, ``Large-scale evolution of image classifiers,'' in
  \emph{Proceedings of the 34th International Conference on Machine
  Learning-Volume 70}.\hskip 1em plus 0.5em minus 0.4em\relax JMLR. org, 2017,
  pp. 2902--2911.

\bibitem{chen2018searching}
L.-C. Chen, M.~Collins, Y.~Zhu, G.~Papandreou, B.~Zoph, F.~Schroff, H.~Adam,
  and J.~Shlens, ``Searching for efficient multi-scale architectures for dense
  image prediction,'' in \emph{Advances in Neural Information Processing
  Systems}, 2018, pp. 8699--8710.

\bibitem{DBLP:journals/corr/abs-1904-07392}
\BIBentryALTinterwordspacing
G.~Ghiasi, T.~Lin, R.~Pang, and Q.~V. Le, ``{NAS-FPN:} learning scalable
  feature pyramid architecture for object detection,'' \emph{CoRR}, vol.
  abs/1904.07392, 2019. [Online]. Available:
  \url{http://arxiv.org/abs/1904.07392}
\BIBentrySTDinterwordspacing

\bibitem{yan2018deeplesion}
K.~Yan, X.~Wang, L.~Lu, and R.~M. Summers, ``Deeplesion: automated mining of
  large-scale lesion annotations and universal lesion detection with deep
  learning,'' \emph{Journal of Medical Imaging}, vol.~5, no.~3, p. 036501,
  2018.

\bibitem{du2018real}
T.~Du, X.~Liu, H.~Zhang, and B.~Xu, ``Real-time lesion detection of cardiac
  coronary artery using deep neural networks,'' in \emph{2018 International
  Conference on Network Infrastructure and Digital Content (IC-NIDC)}.\hskip
  1em plus 0.5em minus 0.4em\relax IEEE, 2018, pp. 150--154.

\bibitem{fang2017object}
Y.~Fang, K.~Kuan, J.~Lin, C.~Tan, and V.~Chandrasekhar, ``Object detection
  meets knowledge graphs,'' 2017.

\bibitem{scherer2010evaluation}
D.~Scherer, A.~M{\"u}ller, and S.~Behnke, ``Evaluation of pooling operations in
  convolutional architectures for object recognition,'' in \emph{International
  conference on artificial neural networks}.\hskip 1em plus 0.5em minus
  0.4em\relax Springer, 2010, pp. 92--101.

\bibitem{gao2019res2net}
S.-H. Gao, M.-M. Cheng, K.~Zhao, X.-Y. Zhang, M.-H. Yang, and P.~Torr,
  ``Res2net: A new multi-scale backbone architecture,'' \emph{arXiv preprint
  arXiv:1904.01169}, 2019.

\bibitem{kipf2017semi}
T.~N. Kipf and M.~Welling, ``Semi-supervised classification with graph
  convolutional networks,'' in \emph{International Conference on Learning
  Representations (ICLR)}, 2017.

\bibitem{yan2018deep}
K.~Yan, X.~Wang, L.~Lu, L.~Zhang, A.~P. Harrison, M.~Bagheri, and R.~M.
  Summers, ``Deep lesion graphs in the wild: relationship learning and
  organization of significant radiology image findings in a diverse large-scale
  lesion database,'' in \emph{Proceedings of the IEEE Conference on Computer
  Vision and Pattern Recognition}, 2018, pp. 9261--9270.

\bibitem{DBLP:journals/corr/LinMBHPRDZ14}
\BIBentryALTinterwordspacing
T.~Lin, M.~Maire, S.~J. Belongie, L.~D. Bourdev, R.~B. Girshick, J.~Hays,
  P.~Perona, D.~Ramanan, P.~Doll{\'{a}}r, and C.~L. Zitnick, ``Microsoft
  {COCO:} common objects in context,'' \emph{CoRR}, vol. abs/1405.0312, 2014.
  [Online]. Available: \url{http://arxiv.org/abs/1405.0312}
\BIBentrySTDinterwordspacing

\bibitem{paszke2017automatic}
A.~Paszke, S.~Gross, S.~Chintala, G.~Chanan, E.~Yang, Z.~DeVito, Z.~Lin,
  A.~Desmaison, L.~Antiga, and A.~Lerer, ``Automatic differentiation in
  pytorch,'' in \emph{NIPS-W}, 2017.

\bibitem{kingma2014adam}
D.~P. Kingma and J.~Ba, ``Adam: A method for stochastic optimization,''
  \emph{arXiv preprint arXiv:1412.6980}, 2014.

\end{thebibliography}


\end{document}


\title{FSD: Fully-Specialized Detector via Neural Architecture Search Supplementary Material}

\maketitle

\section{Additional Details on Experiments}

\subsection{DeepLesion Dataset}
To demonstrate the effectiveness of our proposed NAS-LesionNet, we carry out extensive experiments on DeepLesion~\cite{yan2018deep} dataset, a representative large-scale lesion detection benchmark dataset containing 32,120 CT key slices with 1-3 lesions annotated per slice, adding up to total 32735 lesions with bounding box. The neighboring slices of those key slices are also provided as optional 3D context information. This dataset is officially randomly split into a training set consisting of 22,496 lesions from 22,919 key slices, a validation set with 4,793 lesions from 4,889 key slices and a test set made up of 4,831 lesions from the rest 4,927 key slices. All the data from official validation and test set are annotated with one of the seven lesion types, which are bone(BN), abdomen(AB), mediastinum(ME), liver(LV), lung(LU), kidney(KD), soft tissue(ST) and pelvis(PV) lesions, respectively, whereas training data are lacking in detailed lesion type information. Thus, the DeepLesion dataset can be used to perform binary lesion detection task as well as multi-type lesion detection study thanks to the great variety of lesion types and provided annotations.

\subsection{Evaluation Criteria}
For a comprehensive evaluation of the detection result of our NAS-LesionNet, we adopt some common metrics in general object detection tasks, which are the mean average precision (mAP) and the overall mean recall rate. Moreover, following~\cite{yan20183d}, we also evaluate our method using the sensitivity under different false positives per image (FPPI) corresponding with different IoU thresholds and their intersection over the detected bounding-box (IoBB) thresholds counterparts. The detailed descriptions are as follows.

\textbf{mAP \& mean recall.} We report MS COCO~\cite{DBLP:journals/corr/LinMBHPRDZ14} style mAP and recall metrics, which are averaged over different IoU thresholds from 0.5 to 0.95 with incremental step size 0.05. We strictly follow its evaluation algorithm as detailed documented at http://cocodataset.org/.

\textbf{Sensitivity at different FPPI.} As is discussed in~\cite{yan20183d}, analyzing sensitivity at different FPPI thresholds is appropriate since there are missing annotations in the test set of the DeepLesion dataset. Besides, corresponding IoBB metrics are also reasonable and pragmetic as detection results whose IoU ratios are low but IoBB ratios are high are still very useful and ideal for providing medical assistant during real-world applications. Thus, we take these task-specific criteria into our consideration as well.

\subsection{Implementation Details}
We first carry out our proposed neural architecture search pipeline (NAS-Lesion) to search for a domian-specific network for lesion detection. Then we apply the network resulted from NAS-Lesion, denoted as NAS-LesionNet, to two different medical detection tasks on the DeepLesion dataset, which are namely the binary lesion detection and multi-type lesion detection with our proposed graph module, denoted as LesionNet w/ Graph. Corresponding implementation details using PyTorch v0.4.1~\cite{paszke2017automatic} are as follows.

\textbf{DeepLesion Data Preprocessing.} Unlike datasets made up of natural images, CT slices from the DeepLesion dataset are formatted as single channel, 16-bit PNG images. For each key slice, we concatenate neighboring 2 slices to form a 3 channel image and follow the official guideline documented at https://nihcc.app.box.com/v/DeepLesion/file/306055882594 for data conversion, resulting in an 8-bit, 3 channel image with RGB values ranged within 0-255.

\textbf{NAS-Lesion Pipeline.} When performing NAS-Lesion, we both search for a detection backbone (referred as AutoBone) suitable for effective and efficient feature representation of medical CT slices and a convolutional head (referred as AutoHead), replacing the heavily-used FC head originally designed for FPN~\cite{lin2017feature}, which contains much more parameters. We adopt second order DARTS~\cite{liu2018darts} as our differentiable NAS algorithm. When searching for the backbone, we keep the first two layers of ResNet-50~\cite{he2015deep} with corresponding Image-Net~\cite{alexnet} pretrained weights and replace the rest with 3 reduction cells. FPN~\cite{lin2017feature} structure is used in all experiments. Following~\cite{liu2018darts}, we randomly split the training set into two halves with equal number of slices each, one for optimizing the weights and another half for optimizing the architecture. We search for a task-specific head using our searched backbone with weights pretrained on binary lesion detection task as backbone and replacing the original FC head with 2 differentiable NAS cells, which is a normal cell followed by a reduction cell. For both NAS process, we set the initial learning rate for network weights optimization to 0.01 per image with SGD optimizer (momentum set to 0.9 and weight decay set to 0.0003) and cosine learning scheduler annealing to 0.0001 gradually. The learning rate for the architecture is set to be 0.0024 per image with Adam~\cite{kingma2014adam} optimizer with 0.001 weight decay. Different from original DARTS~\cite{liu2018darts}, we do not have stem blocks since none of our two architecture optimizations involve raw input layers. Also, we do not utilize path dropout to avoid making two-stage detection training unstable. We augment our search results as suggested by~\cite{liu2018darts} and augment details are be described in following sections.

\textbf{LesionNet for Binary Lesion Detection.} After NAS-Lesion searching with our proposed search spaces specialized for medical data, following~\cite{liu2018darts}, we formulate our backbone(AutoBone) and head(AutoHead) by augmenting based on searched results. Empirically, we augment our AutoBone by adding 3 normal cells before each reduction cell, with all normal cells sharing the same structure of reduction cell for simplicity. Similarly, we augment our AutoHead by adding 2 normal cells before the normal cell and reduction cell originated from searching stage, together to form a 4-layer convolutional head. We then exploit the fully-automated task-specified detection network to perform binary lesion detection of the DeepLesion dataset from scratch, retrieving state-of-the-art results.

\textbf{LesionNet w/ Graph for Multi-class Lesion Detection.} Unlike other datasets, the DeepLesion dataset provides researchers with multi-type lesion annotations. For multi-class lesion detection task, we make use of a graph reasoning strategy, extending the existing detection head by adding our proposed graph module after the 9-class classifier, resulting in improvement of performance on multi-class lesion detection and interpretable visualization result to explain the effectiveness of our design. Since only the validation and the test set have lesion subtype information, we use the official validation set as training set and test on the test set. Due to the small amount of total annotated lesions for multi-class detection (only 30\% of the DeepLesion dataset), we finetune the binary lesion detection networks instead of training multi-class detection from scratch.

\textbf{Detailed Settings for Detection.} For all of our detection experiments (including binary and multi-class lesion detection mentioned above), we adopt Feature Pyramid Network (FPN)~\cite{lin2017feature} structure, including our baseline, as it consistently benefits the performance of various kind of detection tasks. In all experiments, the shorter size of input image is set to 512. Anchor ratios are 0.5, 1, 2 and anchor scales are 2, 3, 4, 6, 12. For binary lesion detection task, we set initial learning rate to be 0.005 per sample and train our networks for 12 epochs with one additional warmup epoch at the very beginning of the training. The learning rate is multiplied by 0.1 when the training reaches the 8th and the 11th epoch and SGD optimizer with momentum 0.9 is used. For multi-class lesion detection task, we set initial learning rate to be 0.00125 per sample to stabilize the finetuning process and finetune for 15 epochs with two additional warmup epochs, using the same learning rate scheduler in aforementioned binary detection. For both binary and multi-class lesion detection, we report the performance on the official test set of the DeepLesion dataset.

\section{Searching Results}
\begin{figure*}
    \centering
    \includegraphics[width=1.0\linewidth]{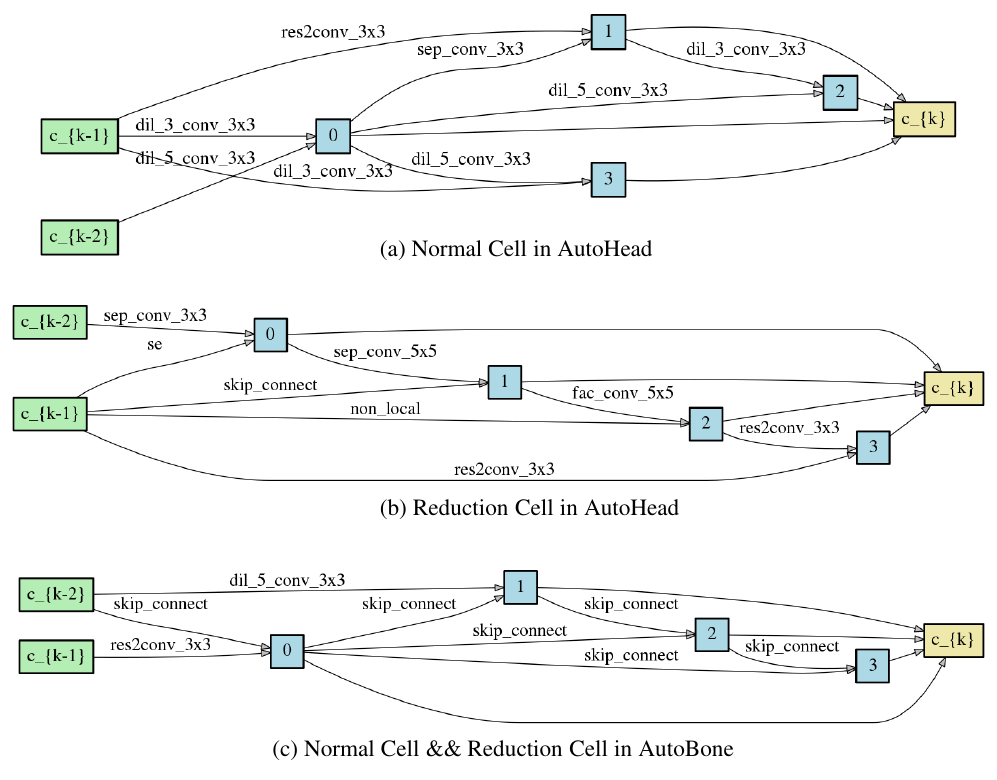}
    \caption{The searching results of our AutoBone and AutoHead.}
    \label{fig:cell}
\end{figure*}

As shown in Fig.~\ref{fig:cell}.

\section{Two-stage Differentiable NAS}

The main purpose of NAS on lesion detection (referred as NAS-Lesion) is to help the design of a network customized and specialized for medical lesion detection. State-of-the-art NAS methods are typically in differentiable manner, where they predefine an operation pool, as search space $O$, and fixed number of ordered latent feature maps, named nodes, with densely-connected edges, forming a directed acyclic graph (DAG), called \textit{parent graph}. The above-mentioned elements formulate a basic searching unit, called a search cell. Each edge $(i \rightarrow j)$, indicating the unidirectional connection from node $x_{i}$ to node $x_{j}$, represents candidate operations, denoted as ${O}_{i \rightarrow j}$, to be learnt by NAS. Thus, we can represent the intermediate computation between input node $x_{i}$ and output node $x_{j}$ as 

\begin{equation}
x_{j} = \sum_{i<j} {O}_{i \rightarrow j}(x_{i}),
\label{eq:operation}
\end{equation}

We follow DARTS~\cite{liu2018darts} to include \textit{zero} operation here and for each cell we have two inputs from previous cells and one concatenated output as well. When it comes to search space continuity, we relax our search space during searching using softmax. Specifically, we define $o(\cdot)$ as operation function and $\alpha_{i \rightarrow j} \in \big\{ o^{\prime} \in O | \alpha^{o^{\prime}}_{i \rightarrow j} \big\}$ for learnable set of architecture weights. For each single operation $o_{i \rightarrow j} \in O_{i \rightarrow j}$, we will have

\begin{align}
\bar{o}_{i \rightarrow j}(x_{j}) = \sum_{o_{i \rightarrow j} \in O_{i \rightarrow j}} \frac{\exp(\alpha^o_{i \rightarrow j})}{\sum_{o' \in O_{i \rightarrow j}} \exp(\alpha^{o'}_{i \rightarrow j})} o(x_{i}), 
\label{eq:softmax}
\end{align}

where $\bar{o}(\cdot)$ represents the mixed operation. This formula can be interpreted as: from $x_{i}$ to ${x_{j}}$ we perform a mixed operation during searching, over the operation search space $O{i \rightarrow j}$ with each operation $o_{i \rightarrow j}$ weighted by corresponding architecture weight denoted as $\alpha^{o_{i \rightarrow j}}_{i \rightarrow j}$. After searching, $\bar{o}_{i \rightarrow j}$ is replaced by the most important operation $o_{i \rightarrow j}$ where $o_{i \rightarrow j} = \mathrm{argmax}_{o \in O_{i \rightarrow j}} \alpha^{o}_{i \rightarrow j}$. For simplicity, we refer to $\alpha$ and $w$ as the encoder of the network architecture and parameter weights, respectively. At this point, we convert the NAS problem into a joint optimization problem by minimizing the training loss $\mathcal{L}^{train}(\cdot)$ and validation loss $\mathcal{L}^{val}(\cdot)$ to find optimal $\alpha^*$ and $w^*$, which are

\begin{align}
& w^* = \mathrm{argmin}_{w} \mathcal{L}^{train}(w, \alpha), \\
& \alpha^* = \mathrm{argmin}_{\alpha} \mathcal{L}^{val}(\mathrm{argmin}_{w} \mathcal{L}^{train}(w, \alpha), \alpha), 
\end{align}

respectively. For image classification tasks, $w$ can simply be parameters of the whole network and $\alpha$ can simply be the same (i.e. set the operation search space to be the same) for every search cell. 

However, for a non-trivial two-stage detection task, $w$ and $\alpha$ will not be obvious as the network components and the detection optimization are more complicated compared with ordinary classification tasks. For having necessary notations, we separate the train loss $\mathcal{L}^{train}$ and the validation loss $\mathcal{L}^{val}$ into $\mathcal{L}^{train}_{bone}$ and $\mathcal{L}^{train}_{head}$, $\mathcal{L}^{val}_{bone}$ and $\mathcal{L}^{val}_{head}$, respectively, representing the corresponding loss of corresponding part of the network. We further separate parameter weights $w$ into $w_{bone}$ and $w_{head}$, $\alpha$ into $\alpha_{bone}$ and $\alpha_{head}$, denoting parameter weights and the architecture encoding of the backbone and the detection head, respectively. As is mentioned above, simple and intuitive optimization strategy resembling NAS on image classification tasks will be something looks like:

\begin{align}
\min_{\alpha_{bone}, \alpha_{head}} \quad & \mathcal{L}^{val}_{bone}(w^*_{bone}(\alpha_{bone})) + \mathcal{L}^{val}_{head}(w^*_{head}(\alpha_{head})) \label{eq:outer} \\
\text{s.t.} \enskip \quad \quad & w^*_{bone}(\alpha_{bone}) = \mathrm{argmin}_{w_{bone}} (\mathcal{L}^{train}_{bone}(w_{bone}, \alpha_{bone}) + \mathcal{L}^{train}_{head}(w_{bone}, \alpha_{bone})), \label{eq:inner} \\
\quad \quad & w^*_{head}(\alpha_{head}) = \mathrm{argmin}_{w_{head}} \mathcal{L}^{train}_{head}(w_{head}, \alpha_{head}). \label{eq:inner2}
\end{align}

With so many different levels of parameters and loss targets, this is even more nested than its classification counterpart, which is already hard to optimize. As two-stage detection alone is difficult to optimize~\cite{he2018rethinking,ioffe2015batch} and hardware constraints such as GPU memory limit exists, it is prohibitive to directly perform NAS optimization throughout the whole two-stage detection pipeline. Thus, in the Method section of the main paper, we explained and discussed how we specifically accommodate these challenges to conduct NAS on the lesion detection task in a de-nested way of optimization.

\vfill

\bibliographystyle{IEEEtran}
\bibliography{egbib}
